\documentclass{article}

\usepackage{times}
\usepackage{graphbox}
\usepackage{subfig}
\usepackage{float}

\usepackage{natbib}

\usepackage{algorithm}
\usepackage{algorithmic}

\usepackage{amsfonts}
\usepackage{amsmath}

\usepackage{kotex}

\usepackage{hyperref}

\usepackage[accepted]{icml2017}

\definecolor{orange}{RGB}{255,127,0}

\definecolor{pink}{RGB}{255,0,127}

\icmltitlerunning{Learning to Discover Cross-Domain Relations with Generative Adversarial Networks}
\begin{document} 

\twocolumn[
\icmltitle{Learning to Discover Cross-Domain Relations \\ with Generative Adversarial Networks} 



% you can specify symbols, otherwise they are numbered in order
\icmlsetsymbol{}{*}

\begin{icmlauthorlist}
\icmlauthor{Taeksoo Kim}{tbrain}
\icmlauthor{Moonsu Cha}{tbrain}
\icmlauthor{Hyunsoo Kim}{tbrain}
\icmlauthor{Jung Kwon Lee}{tbrain}
\icmlauthor{Jiwon Kim}{tbrain}
\end{icmlauthorlist}

\icmlaffiliation{tbrain}{SK T-Brain, Seoul, South Korea}
%\icmlaffiliation{usc}{Viterbi Department of Computer Science, USC, Los Angeles, CA.}

\icmlcorrespondingauthor{Taeksoo Kim}{jazzsaxmafia@sktbrain.com}
%\icmlcorrespondingauthor{Eee Pppp}{ep@eden.co.uk}

\icmlkeywords{Cross-Domain Relation Discovery, Generative Adversarial Network, Image-to-Image Translation}

\vskip 0.3in 
]



\printAffiliationsAndNotice{}  % leave blank if no need to mention equal contribution
%\footnotetext{hi}

% --------------------------------------------------------------------------------------------------------
% --------------------------------------------- A B S T R A C T ------------------------------------------
% --------------------------------------------------------------------------------------------------------
\begin{abstract}
While humans easily recognize relations between data from different domains without any supervision, learning to automatically discover them is in general very challenging and needs many ground-truth pairs that illustrate the relations. To avoid costly pairing, we address the task of discovering cross-domain relations given unpaired data. We propose a method based on generative adversarial networks that learns to discover relations between different domains (DiscoGAN). Using the discovered relations, our proposed network successfully transfers style from one domain to another while preserving key attributes such as orientation and face identity. 
\end{abstract}

\newcommand{\ie}{i.e.\ }
\newcommand{\eg}{e.g.\ }
\newcommand{\figref}[1]{Figure \ref{fig:#1}}

\section{Introduction}
\label{Introduction}

%A picture is worth thousand words. Representation based on images story.. but maybe time is too short to write this story. 

% Let's imagine yourself thinking about what to wear today... => suit jacket vs t-shirt?
Relations between two different domains, the way in which concepts, objects, or people are connected, arise ubiquitously.
Cross-domain relations are often natural to humans.
For example, we recognize the relationship between an English sentence and its translated sentence in French. We also choose a suit jacket with pants or shoes in the same style to wear.
%- Cross-domain relations are often natural to humans. When two same-meaning sentences in different languages are given to a person speaking both languages, he or she can easily see that two sentences mean the same. Similarly, most people are able to figure out if a sketch and a photo indicate the same person. This is possible partially due to that humans have learned and exploited existing knowledge of concepts, objects and people.     
%- Machine learning models, however, are mostly trained from scratch and models can only learn knowledge available within a given dataset. When ground-truth data relating cross-domain data samples are given, supervised learning can be used to model relations. Human labeling indicating which data samples are related to which can be very costly.    

Can machines also achieve a similar ability to relate two different image domains?
%, and further generating a corresponding image of one domain given an image from another domain?
This question can be reformulated as a conditional image generation problem. In other words, finding a mapping function from one domain to the other can be thought as generating an image in one domain given another image in the other domain.
While this problem tackled by generative adversarial networks (GAN) \cite{isola2016image} has gained a huge attention recently, most of today's training approaches use explicitly paired data, provided by human or another algorithm.

%\hspace*{-5mm}
\begin{figure}[t!]
  %\vskip 0.2in
 \small
 \begin{center}
  \hspace*{-8mm}
  \begin{tabular}[!htbp]{ c }
  \includegraphics[width=1.03\columnwidth]{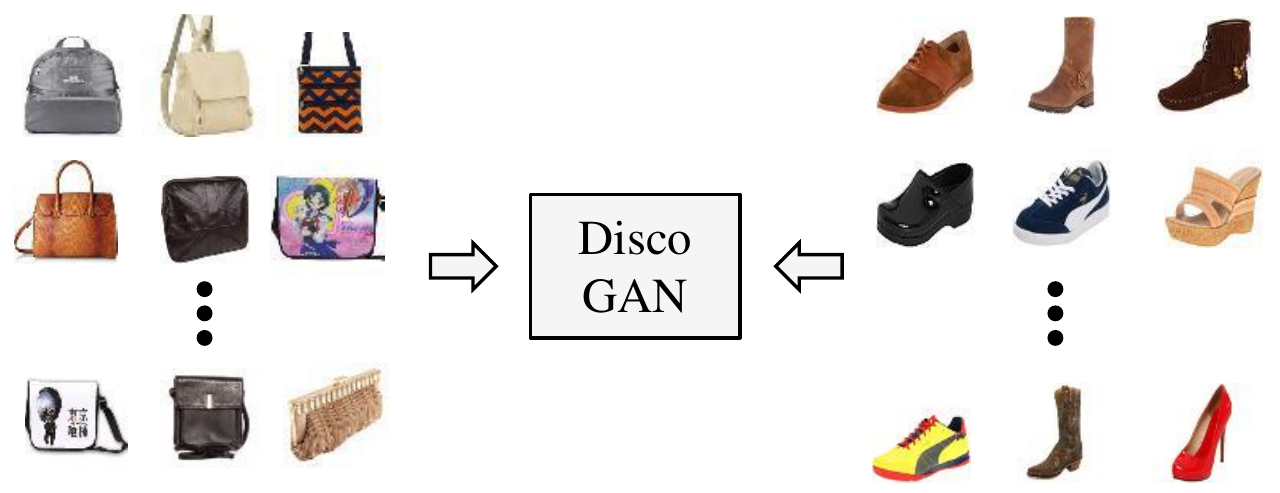}\\
  (a) Learning cross-domain relations \textbf{without any extra label}\\ 
  \includegraphics[width=1.09\columnwidth]{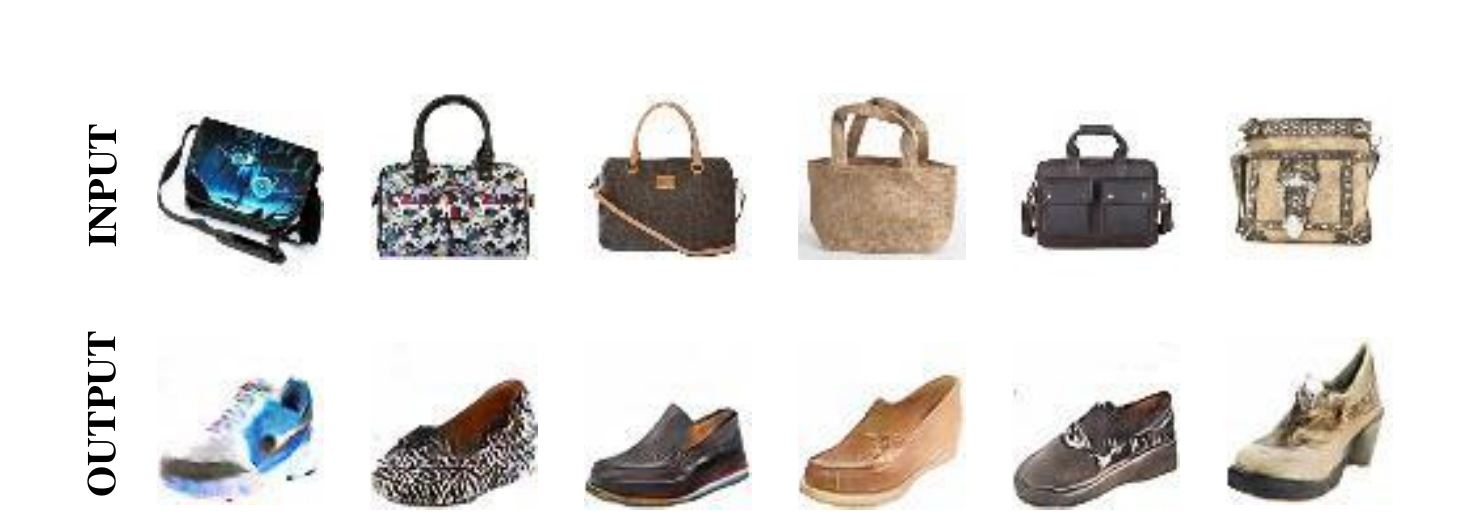}\\
  (b) Handbag images (input) \hspace{2px} \& \hspace{2px} \textbf{Generated} shoe images (output)\\
  \includegraphics[width=1.09\columnwidth]{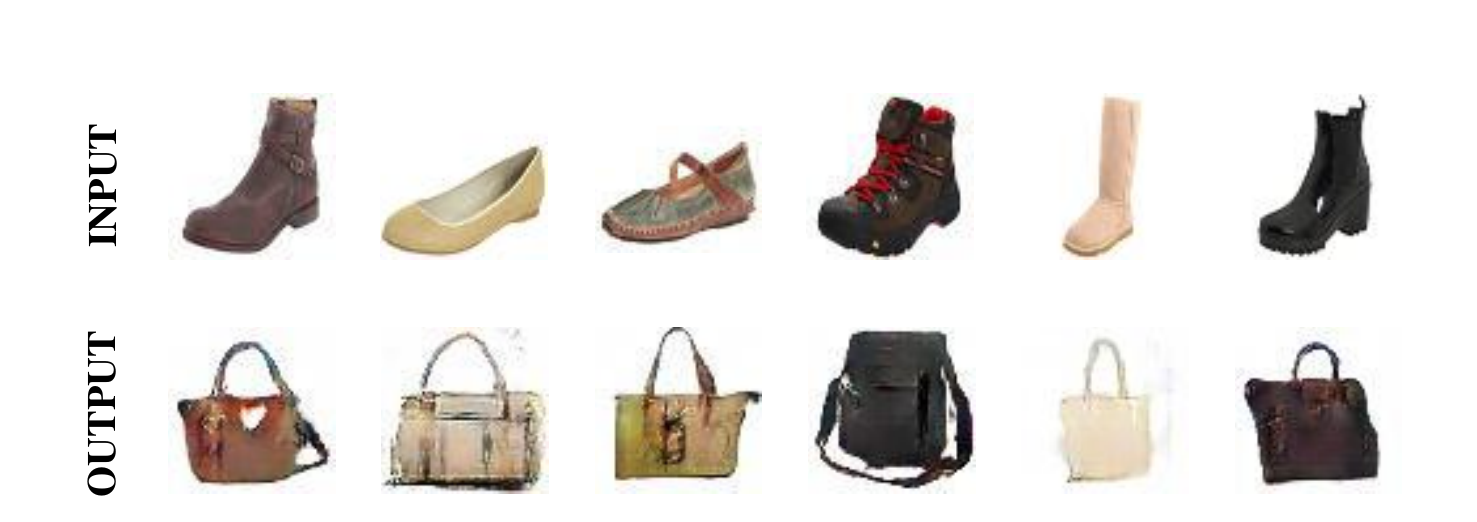}\\
  (c) Shoe images (input) \hspace{2px} \& \hspace{2px} \textbf{Generated} handbag images (output) \\
  \end{tabular}
  
  \caption{Our GAN-based model trains with two independently collected sets of images and learns how to map two domains \textit{without any extra label}. In this paper, we reduces this problem into generating a new image of one domain given an image from the other domain.
  (a) shows a high-level overview of the training procedure of our model with two independent sets (\eg handbag images and shoe images).
  (b) and (c) show results of our method. Our method takes a handbag (or shoe) image as an input, and generates its corresponding shoe (or handbag) image.
  Again, it's worth noting that our method does not take any extra annotated supervision and can self-discover relations between domains.
  }
  \label{fig:teaser}
  
  \end{center}
  \vskip -0.2in
\end{figure}

\begin{figure*}[htbp]
  \begin{center}
  \centerline{\includegraphics[width=\textwidth]{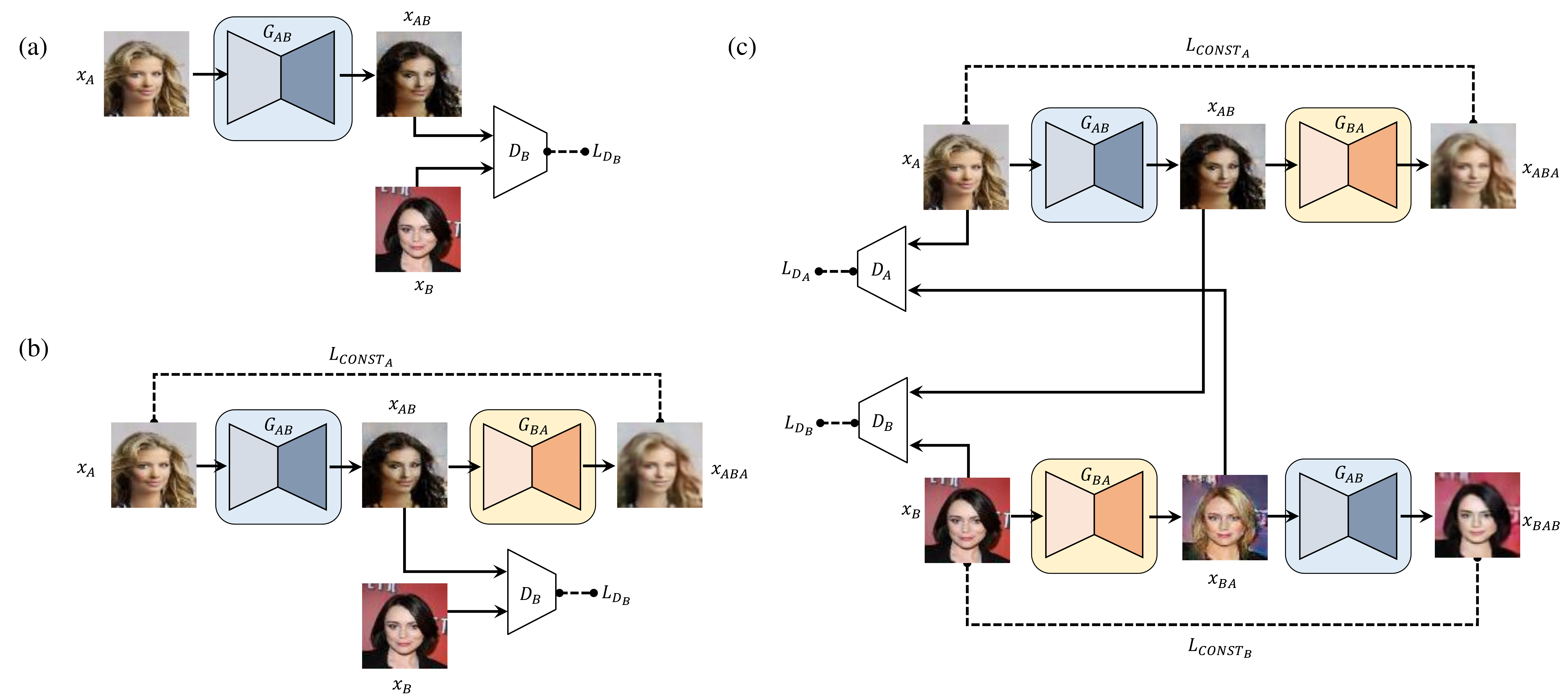}}
  \caption{Three investigated models. (a) standard GAN \cite{goodfellow2014generative}, (b) GAN with a reconstruction loss, (c) our \textbf{proposed model (DiscoGAN)} designed to discover relations between two unpaired, unlabeled datasets. Details are described in Section 3.}
\label{fig:architecture}
  \end{center}
  \vskip -0.3in
\end{figure*}

This problem also brings an interesting challenge from a learning point of view.
Explicitly supervised data is seldom available and labeling can be labor intensive. Moreover, pairing images can become tricky if corresponding images are missing in one domain or there are multiple best candidates.
Hence, we push one step further by discovering relations between two visual domains without any explicitly paired data.
%Hence, we push one step further by  a relationship between two visual domains without any explicitly paired data.
%In other words, how can we learn a generative transferring function from one visual domain to another visual domain?

In order to tackle this challenge, we introduce a model that discovers cross-domain relations with GANs (DiscoGAN).
Unlike previous methods, our model can be trained with two sets of images without any explicit pair labels (see \figref{teaser}a) and does not require any pre-training.
Our proposed model can then take one image in one domain as an input and generate its corresponding image in another domain (see \figref{teaser}b).
The core of our model is based on two different GANs coupled together -- each of them ensures our generative functions can map each domain to its counterpart domain.
A key intuition we rely on is to constraint \textit{all images} in one domain to be \textit{representable} by images in the other domain. 
%This simply means that each image in domain A can generate a new image in domain B and hence represented by this generated image; and this generated image can then reconstruct back the original image.
For example, when learning to generate a shoe image based on each handbag image, we force this generated image to be an image-based representation of the handbag image (and hence reconstruct the handbag image) through a reconstruction loss, and to be as close to images in the shoe domain as possible through a GAN loss.
%an image of a handbag generates a shoe image of the similar style, and we force this generated shoe image as a latent image-based representation of the handbag and hence can reconstruct the handbag image.
We use these two properties to encourage the mapping between two domains to be well covered on both directions (\ie encouraging one-to-one rather than many-to-one or one-to-many).
In the experimental section, we show that this simple intuition discovered common properties and styles of two domains very well.

%In short, we exploit the fact that all images in one domain need to be representable by another domain.

%discover an underlying relationship between domains by attempting to maximize the mapping coverage from each other.
%Our model encourages generating the best representative image in the other domain for each image in one domain, and also 
%Our key intuition is to learn a generative network that can generate the best representative image in domain $B$ given an image in domain $A$. This representative image is defined by (1) representativeness: how well it can be used to reconstruct the original image, and (2) 
%Our key intuition is to train a generative network that can generate the best pairing image, which can well reconstruct back the original image.

% - Our model called Disco GAN can generate images by finding core relations between image domains.
% Two key insights of our model are:
% - Our model can learn with pairless data.
% - (1) Keyword1 (representation?): By learning a function $G_{AB}$ generating the best image $x_B$ to preserves the information of $x_A$. (2) Keyword2 (): But, by (1) alone, it can start ``cheating'' (storing a small code of $x_A$ on the side so that $G_{BA}$ can pick it up). So, we add more constraints to $G_{BA}$ by the double structure.

Both experiments on toy domain and real world image datasets support the claim that our proposed model is well-suited for discovering cross-domain relations. When translating data points between simple 2-dimensional domains and between face image domains, our DiscoGAN model was more robust to the mode collapse problem compared to two other baseline models. It also learns the bidirectional mapping between two image domains, such as faces, cars, chairs, edges and photos, and successfully apply them in image translation. Translated images consistently change specified attributes such as hair color, gender and orientation while maintaining all other components. Results also show that our model is robust to repeated application of translation mappings.

\begin{figure*}[!h]
  %\vskip 0.2in
  \begin{center}
  \centerline{\includegraphics[width=\textwidth]{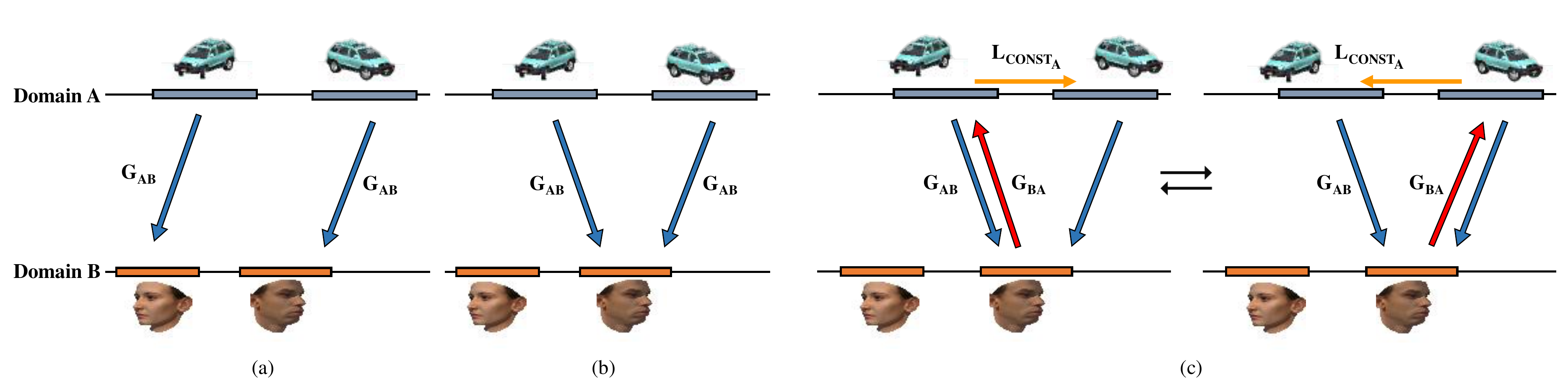}}
  \caption{Illustration of our models on simplified one dimensional domains. (a) ideal mapping from domain A to domain B in which the two domain A modes map to two different domain B modes, (b) GAN model failure case, (c) GAN with reconstruction model failure case.}
  \label{fig:1d_illustration}
  \end{center}
  \vskip -0.2in
\end{figure*} 
  
\section{Model}
\label{Model}
We now formally define cross-domain relations and present the problem of learning to discover such relations in two different domains. Standard GAN model and a similar variant model with additional components are investigated for their applicability for this task. Limitations of these models are then explained, and we propose a new architecture based on GANs that can be used to discover cross-domain relations.

\subsection{Formulation} \label{sec:formulation}
Relation is mathematically defined as a function $\textbf{G}_{AB}$ that maps elements from its domain $A$ to elements in its codomain $B$ and $\textbf{G}_{BA}$ is similarly defined. In fully unsupervised setting, $\textbf{G}_{AB}$ and $\textbf{G}_{BA}$ can be arbitrarily defined. To find a meaningful relation, we need to impose a condition on the relation of interest. Here, we constrain relation to be a one-to-one correspondence (bijective mapping). That means $\textbf{G}_{AB}$ is the inverse mapping of $\textbf{G}_{BA}$. 

The range of function $\textbf{G}_{AB}$, the complete set of all possible resulting values $\textbf{G}_{AB}(x_A)$ for all $x_A$'s in domain $A$, should be contained in domain $B$ and similarly for $\textbf{G}_{BA}(x_B)$. 

We now relate these constraints to objective functions. Ideally, the equality $\textbf{G}_{BA}\circ \textbf{G}_{AB}(x_A) = x_A$ should be satisfied, but this hard constraint is difficult to optimize and relaxed soft constraint is more desirable in the view of optimization. For this reason, we minimize the distance $d(\textbf{G}_{BA}\circ \textbf{G}_{AB}(x_A), x_A)$, where any form of metric function ($L_1$, $L_2$, Huber loss) can be used. Similarly, we also need to minimize $d(\textbf{G}_{AB}\circ \textbf{G}_{BA}(x_B), x_B)$.

Guaranteeing that $\textbf{G}_{AB}$ maps to domain $B$ is also very difficult to optimize. We relax this constraint as follows: we instead minimize generative adversarial loss $-\mathbb{E}_{x_{\!_A} \sim P_{\!_A}} \left[\log \textbf{D}_B(\textbf{G}_{AB}(x_{\!_A}))\right]$. Similarly, we minimize $-\mathbb{E}_{x_{\!_B} \sim P_{\!_B}} \left[\log \textbf{D}_A(\textbf{G}_{BA}(x_{\!_B}))\right]$.

Now, we explore several GAN architectures to learn with these loss functions.

\subsection{Notation and Architecture}
We use the following notations in sections below. A generator network is denoted $\textbf{G}_{AB}: \mathbb{R}^{64\times64\times3}_{A} \rightarrow \mathbb{R}^{64\times64\times3}_{B}$, and the subscripts denote the input and output domains and superscripts denote the input and output image size. The discriminator network is denoted as $\textbf{D}_B: \mathbb{R}^{64\times64\times3}_B \rightarrow [0,1]$, and the subscript B denotes that it discriminates images in domain B. Notations $\textbf{G}_{BA}$ and $\textbf{D}_A$ are used similarly. 

Each generator takes image of size $64 \times 64 \times 3$ and feeds it through an encoder-decoder pair. The encoder part of each generator is composed of convolution layers with $4 \times 4$ filters, each followed by leaky ReLU \cite{maas2013leakyrelu, xu2015rectifier}. The decoder part is composed of deconvolution layers with $4 \times 4$ filters, followed by a ReLU, and outputs a target domain image of size $64 \times 64 \times 3$. The number of convolution and deconvolution layers ranges from four to five, depending on the domain. 

The discriminator is similar to the encoder part of the generator. In addition to the convolution layers and leaky ReLUs, the discriminator has an additional convolution layer with  $4 \times 4$ filters, and a final sigmoid to output a scalar output between $[0,1]$.  
  
% \begin{figure}[!htbp]
%   %\vskip 0.2in
%   \begin{center}
%   \centerline{\includegraphics[width=\columnwidth]{Figures/illustration_figure_2_v0_0_1_2017_02_24}}
%   \caption{Illustration of our proposed model}
%   \label{fig:1d_illustration_2}
%   \end{center} 
%   \vskip -0.2in
%   \end{figure} 

\subsection{GAN with a Reconstruction Loss}
%\textbf{reference + explain (input is image, rather than z noise)}
We first consider a standard GAN model \cite{goodfellow2014generative} for the relation discovery task (Figure \ref{fig:architecture}a). Originally, a standard GAN takes random Gaussian noise $z$, encodes it into hidden features $h$ and generates images such as MNIST digits. We make a slight modification to this model to fit our task: the model we use takes in image as input instead of noise.

In addition, since this architecture only learns one mapping from domain A to domain B, we add a second generator that maps domain B back into domain A (Figure \ref{fig:architecture}b). We also add a reconstruction loss term that compares the input image with the reconstructed image. With these additional changes, each generator in the model can learn mapping from its input domain to output domain and discover relations between them.  

%\textbf{architecture and equation}\\
A generator $\textbf{G}_{AB}$ translates input image $x_{\!_A}$ from domain A into $x_{\!_{AB}}$ in domain B. The generated image is then translated into a domain A image $x_{\!_{ABA}}$ to match the original input image (Equation \ref{eq:gan_1}, \ref{eq:gan_2}). Various forms of distance functions, such as MSE, cosine distance, and hinge-loss, can be used as the reconstruction loss $d$ (Equation \ref{eq:gan_3}). The translated output $x_{\!_{AB}}$ is then scored by the discriminator which compares it to a real domain B sample $x_{\!_B}$.
    \begin{align}
          x_{\!_{AB}} &= \textbf{G}_{AB}(x_{\!_A}) \label{eq:gan_1} \\
          x_{\!_{ABA}} &= \textbf{G}_{BA}(x_{\!_{AB}}) = \textbf{G}_{BA}\circ \textbf{G}_{AB}(x_{\!_A}) \label{eq:gan_2} \\     
          L_{\!_{CONST_A}} &= d(\textbf{G}_{BA} \circ \textbf{G}_{AB} (x_{\!_A}), x_{\!_A}) \label{eq:gan_3}  \\       
          L_{\!_{GAN_B}} &= -\mathbb{E}_{x_{\!_A} \sim P_{\!_A}} \left[\log \textbf{D}_{B}(\textbf{G}_{AB}(x_{\!_A}))\right] \label{eq:gan_4} 
    \end{align}

The generator $\textbf{G}_{AB}$ receives two types of losses -- a reconstruction loss $L_{\!_{CONST_A}}$ (Equation \ref{eq:gan_3}) that measures how well the original input is reconstructed after a sequence of two generations, and a standard GAN generator loss $L_{\!_{GAN_B}}$ (Equation \ref{eq:gan_4}) that measures how realistic the generated image is in domain B. The discriminator receives the standard GAN discriminator loss of Equation \ref{eq:gan_6}. %Again, $L_{CONST_A}$ computes how well the reconstructed output $x_{\!_{ABA}}$ from two sequential generators matches the original input $x_{\!_A}$, while $L_{GAN_A}$ is a standard generator loss for GANs. 
\begin{align}
            L_{\!_{G_{AB}}} &= L_{\!_{GAN_B}} + L_{\!_{CONST_A}} \label{eq:gan_5} 
    \end{align}
    \begin{eqnarray} 
    %\begin{split} 
          L_{\!_{D_B}} & = & -\ \mathbb{E}_{x_{\!_B} \sim P_{\!_B}} [ \log  \textbf{D}_{B}(x_{\!_B}) ] \nonumber\\
                   &&- \ \mathbb{E}_{x_{\!_A} \sim P_{\!_A}} \left[ \log (1 - \textbf{D}_{B}(\textbf{G}_{AB}(x_{\!_A}))\right)] \label{eq:gan_6} 
    %\end{split} 
    \end{eqnarray}
  
 % . The translated image is then back-translated into a source domain image $x_{\!_{ABA}}$ by the second generator $\textbf{G}_{BA}$ to match the original input image. Various forms of distance functions, such as MSE, cosine distance, and hinge-loss, can be used, as the reconstruction loss $d$ (Equation \ref{eq:gan_recon_2}) 
% \begin{align}
%      L_{\!_{CONST_A}} &= d(\textbf{G}_{BA} \circ \textbf{G}_{AB} (x_{\!_A}), x_{\!_A}) %\label{eq:gan_recon_2}  \\
%      L_{\!_{G}} &= L_{\!_{GAN_A}} + L_{\!_{CONST_A}} \label{eq:gan_recon_3}
% \end{align}

%The generator now receives two types of losses -- a GAN generator loss $L_{GAN_A}$ (Equation \ref{eq:gan_2}) and a reconstruction loss $L_{CONST_A}$ (Equation \ref{eq:gan_recon_2}), and the discriminator receives the same loss as Equation \ref{eq:gan_3}. Again, $L_{CONST_A}$ computes how well the back-translated output $x_{\!_{ABA}}$ from two sequential generators matches the original input $x_{\!_A}$, while $L_{GAN_A}$ is a standard generator loss for GANs. The discriminator is the same as the one in a standard GAN model.

%\textbf{problems}\\
During training, the generator $\textbf{G}_{AB}$ learns the mapping from domain A to domain B under two relaxed constraints: that domain A maps to domain B, and that the mapping on domain B is reconstructed to domain A. However, this model lacks a constraint on mapping from B to A, and these two conditions alone does not guarantee a cross-domain relation (as defined in section \ref{sec:formulation}) because the mapping satisfying these constraints is one-directional. In other words, the mapping is an injection, not bijection, and one-to-one correspondence is not guaranteed. 

%Therefore this process can be viewed as continuous space injection that aligns the two factors axes. Although a mapping from factors in $f_x$ to factors in $f_y$ can be learned, the injection mechanism does not guarantee a one-to-one match between the two. As such, multiple values disentangling factors of domain A can be matched with one value in domain B. 

%In addition to this injection mechanism in the latent domain, 

Consider the two possibly multi-modal image domains A and B. Figure \ref{fig:1d_illustration} illustrates the two multi-modal data domains on a simplified one-dimensional representation. Figure \ref{fig:1d_illustration}a shows the ideal mapping from input domain A to domain B, where each mode of data is mapped to a separate mode in the target domain. Figure \ref{fig:1d_illustration}b, in contrast, shows the mode collapse problem, a prevalent phenomenon in GANs, where data from multiple modes of a domain map to a single mode of a different domain. For instance, this case is where the mapping $\textbf{G}_{AB}$ maps images of cars in two different orientations into the same mode of face images. 

In some sense, the addition of a reconstruction loss to a standard GAN is an attempt to remedy the mode collapse problem. In Figure \ref{fig:1d_illustration}c, two domain A modes are matched with the same domain B mode, but the domain B mode can only direct to one of the two domain A modes. Although the additional reconstruction loss $L_{\!_{CONST_A}}$ forces the reconstructed sample to match the original (Figure \ref{fig:1d_illustration}c), this change only leads to a similar symmetric problem. The reconstruction loss leads to an oscillation between the two states and does not resolve mode-collapsing.

\subsection{Our Proposed Model: Discovery GAN}
%\textbf{approach / architecture}
Our proposed GAN model for relation discovery -- DiscoGAN -- couples the previously proposed model (Figure \ref{fig:architecture}c). Each of the two coupled models learns the mapping from one domain to another, and also the reverse mapping to for reconstruction. The two models are trained together simultaneously. The two generators $\textbf{G}_{AB}$'s and the two generators $\textbf{G}_{BA}$'s share parameters, and the generated images $x_{\!_{BA}}$ and $x_{\!_{AB}}$ are each fed into separate discriminators $L_{D_A}$ and $L_{D_B}$, respectively. 

%\textbf{equation}\\
One key difference from the previous model is that input images from both domains are reconstructed and that there are two reconstruction losses: $L_{CONST_A}$ and $L_{CONST_B}$. 
 \begin{align}
      L_{\!_{G}} &= L_{\!_{G_{AB}}} + L_{\!_{G_{BA}}} \label{eq:discogan_1}\\
      			 &= L_{\!_{GAN_B}} + L_{\!_{CONST_A}} + L_{\!_{GAN_A}} + L_{\!_{CONST_B}} \nonumber \\
                 \nonumber \\
      L_{\!_{D}} &= L_{\!_{D_{A}}} + L_{\!_{D_{B}}} \label{eq:discogan_2}
\end{align}

As a result of coupling two models, the total generator loss is the sum of GAN loss and reconstruction loss for each partial model (Equation \ref{eq:discogan_1}). Similarly, the total discriminator loss $L_D$ is a sum of discriminator loss for the two discriminators $\textbf{D}_A$ and $\textbf{D}_B$, which discriminate real and fake images of domain A and domain B (Equation \ref{eq:discogan_2}).

%\textbf{problem solved}
Now, this model is constrained by two $L_{GAN}$ losses and two $L_{CONST}$ losses. Therefore a bijective mapping is achieved, and a one-to-one correspondence, which we defined as cross-domain relation, can be discovered.

  %\hsk{TODO}

% ------------------------------------------------------------------------------------------------------
% --------------------------------------------- EXPERIMENT ---------------------------------------------
% ------------------------------------------------------------------------------------------------------

%  \begin{figure*}[!ht]
%   %\vskip 0.2in
%   \begin{center}
%   \centerline{\includegraphics[width=\textwidth]{Figures/pairGanFigure_toy_small}}
%   \caption{Toy domain experiment}
%   \label{fig:toy}
%   \end{center}
%   \vskip -0.2in
%   \end{figure*} 

\begin{figure*}[!h]
  %\vskip 0.2in
 \small
 \begin{center}
  \setlength{\tabcolsep}{1pt}
  \begin{tabular}[!htbp]{ c c c c }
  \includegraphics[height=4.1cm]{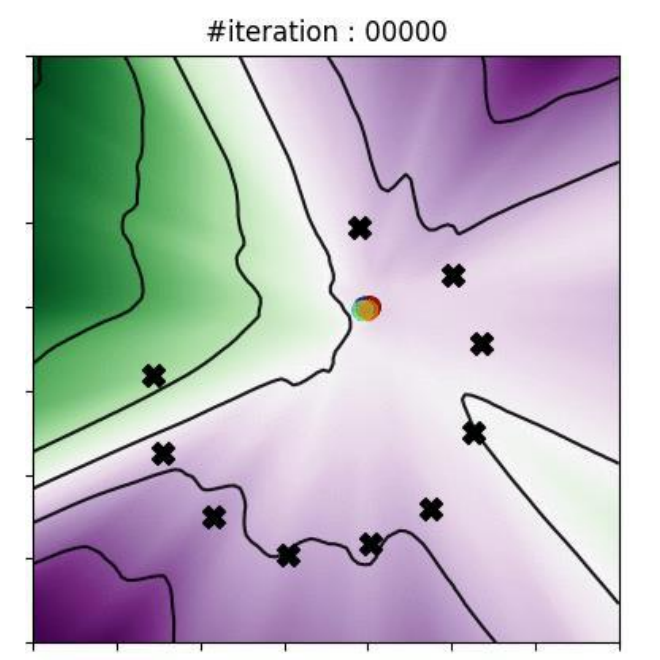} 
  & \includegraphics[height=4.1cm]{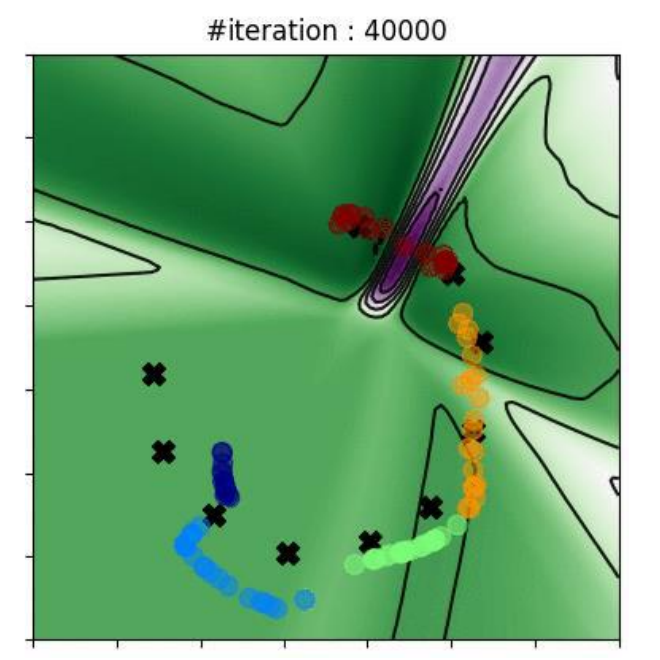} 
  & \includegraphics[height=4.1cm]{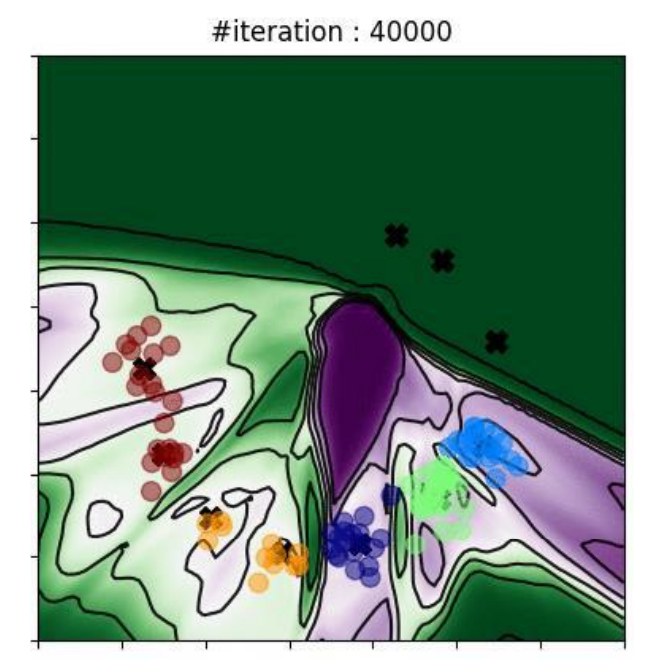} 
  & \includegraphics[height=4.1cm]{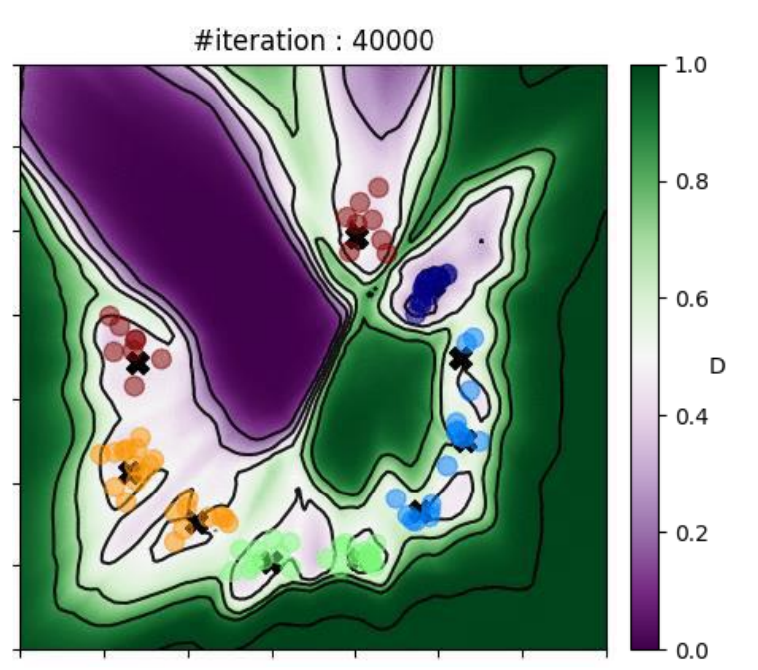} \\
  \small{(a)}
  & \small{(b)}
  & \small{(c)}
  & \small{(d) ~~~~~ } \\
  \end{tabular}
  
  \caption{Toy domain experiment results. The colored background shows the output value of the discriminator. 'x' marks denote different modes in B domain, and colored circles indicate mapped samples of domain A to domain B, where each color corresponds to a different mode. (a) ten target domain modes and initial translations, (b) standard GAN model, (c) GAN with reconstruction loss, (d) our proposed model DiscoGAN}
  \label{fig:toy}
  
  \end{center}
  \vskip -0.2in
\end{figure*}

\section{Experiments}
\subsection{Toy Experiment}
To empirically demonstrate our explanations on the differences between a standard GAN, a GAN with reconstruction loss and our proposed model (DiscoGAN), we designed an illustrative experiment based on synthetic data in 2-dimensional A and B domains. Both source and target data samples are drawn from Gaussian mixture models.

In Figure \ref{fig:toy}, the left-most figure shows the initial state of toy experiment where all the A domain modes map to almost a single point because of initialization of the generator. For all other plots the target domain 2D plane is shown with target domain modes marked with black `x's. Colored points on B domain planes represent samples from A domain that are mapped to the B domain, and each color denotes samples from each A domain mode. In this case, the task is to discover cross-domain relations between the A and B domain and translate samples from five A domain modes into the B domain, which has ten modes spread around the arc of a circle. 

We use a neural network with three linear layers that are each followed by a ReLU nonlinearity as the generator. For the discriminator we use five linear layers that are each followed by a ReLU, except for the last layer which is switched out with a sigmoid that outputs a scalar $\in[0,1]$. The colored background shows the output value of the discriminator $\textbf{D}_B$, which discriminates real target domain samples from synthetic, translated samples from domain A. The contour lines show regions of same discriminator value. 
																																																							
The training was performed for 50,000 iterations, and due to the domain simplicity our model often converged much earlier. The results from this experiment match our claim and illustrations in Figure \ref{fig:toy} and the resulting translated samples show very different behavior depending on the model used. 

In the baseline (standard GAN) case, many translated points of different colors are located around the same B domain mode. For example, navy and light-blue colored points are located together, as well as green and orange colored points. This result illustrates the mode-collapse problem of GANs since points of multiple colors (multiple A domain modes) are mapped to the same B domain mode. The baseline model still oscillate around B modes throughout the iterations.

In the case of GAN with a reconstruction loss, the collapsing problem is less prevalent, but navy, green and light-blue points still overlap at a few modes.  The contour plot also demonstrates the difference from baseline: regions around all B modes are leveled in a green colored plateau in the baseline, allowing translated samples to freely move between modes, whereas in the single model case the regions between B modes are clearly separated. 

In addition, both this model and the standard GAN model fail to cover all modes in B domain since the mapping from A domain to B domain is injective. Our proposed DiscoGAN model, on the other hand, is able to not only prevent mode-collapse by translating into distinct well-bounded regions that do not overlap, but also generate B samples in all ten modes as the mappings in our model is bijective. It is noticeable that the discriminator for B domain is perfectly fooled by translated samples from A domain around B domain modes.

Although this experiment is limited due to its simplicity, the results clearly support the superiority of our proposed model over other variants of GANs.

\begin{figure*}[!h]
  %\vskip 0.2in
  \begin{center}
  \centerline{\includegraphics[width=\textwidth]{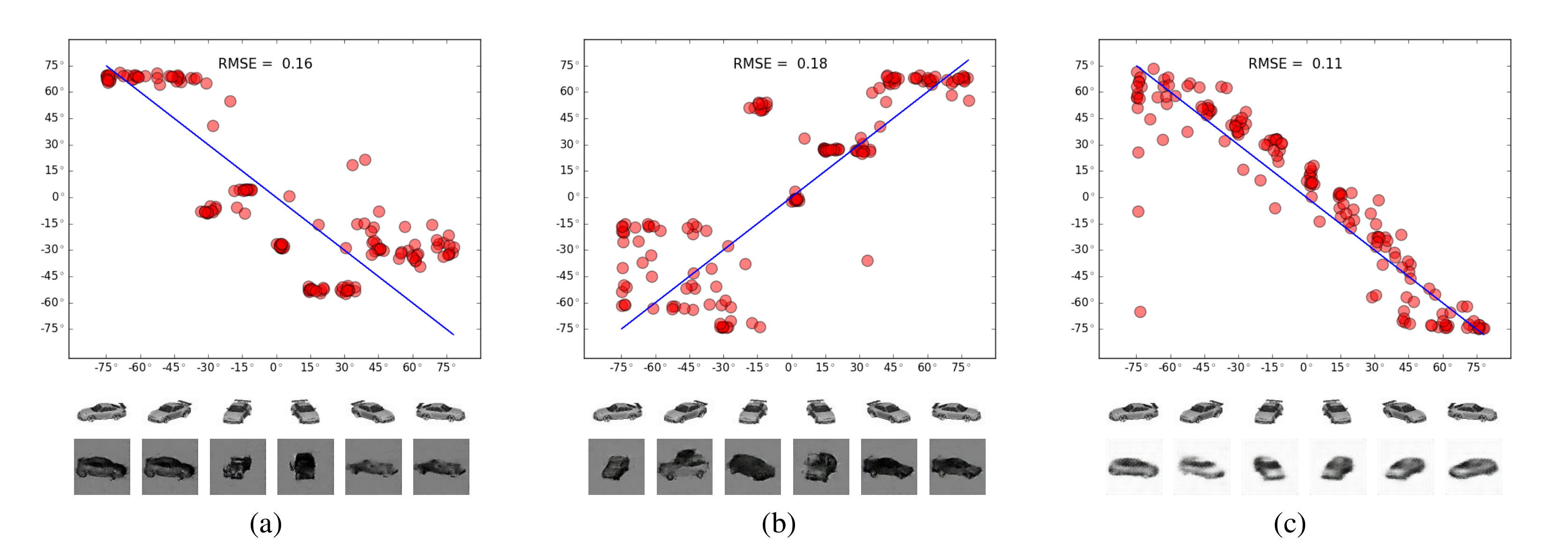}}
  \caption{Car to Car translation experiment. Horizontal and vertical axes in the plots indicate predicted azimuth angles of input and translated images, where the angle of input image ranges from -75$^\circ$ to 75$^\circ$. RMSE with respect to ground truth (blue lines) are shown in each plot. Images in the second row are examples of input car images ranging from -75$^\circ$ to 75$^\circ$ at 15$^\circ$ intervals. Images in the third row are corresponding translated images. (a) plot of standard GAN (b) GAN with reconstruction (c) DiscoGAN. The angles of input and output images are highly correlated when our proposed DiscoGAN model is used. Note the angles of input and translated car images are reversed with respect to 0$^\circ$ (i.e. mirror images).}
  \label{fig:car2car_plot}
  \end{center}
  \vskip -0.2in
  \end{figure*}

\subsection{Real Domain Experiment }
To evaluate whether our DiscoGAN successfully learns underlying relationship between domains, we trained and tested our model using several image-to-image translation tasks that require the use of discovered cross-domain relations between source and target domains. 

In each real domain experiment, all input images and translated images were of size $64 \times 64 \times 3$. For training, we used learning rate of 0.0002 and used the Adam optimizer \cite{kingma15adam} with $\beta_1=0.5$ and $\beta_2 = 0.999$. We applied Batch Normalization \cite{ioffe2015batchnorm} to all convolution and deconvolution layers except the first and the last layers, weight decay regularization coefficient of $10^{-4}$ and minibatch of size 200. All computations were conducted on a single machine with an Nvidia Titan X Pascal GPU and an Intel(R) Xeon(R) E5-1620 CPU.

\subsubsection{Car to Car, Face to Face} 
We used a Car dataset \cite{fidler2012car} which consists of rendered images of 3D car models with varying azimuth angles at 15$^\circ$ intervals. We split the dataset into train and test sets and again split the train set into two groups, each of which is used as A domain and B domain samples. In addition to training a standard GAN model, a GAN with a reconstruction model and a proposed DiscoGAN model, we also trained a regressor that predicts the azimuth angle of a car image using the train set. To evaluate, we translated images in the test set using each of the three trained models, and azimuth angles were predicted using the regressor for both input and translated images. Figure \ref{fig:car2car_plot} shows the predicted azimuth angles of input and translated images for each model. In standard GAN and GAN with reconstruction (\ref{fig:car2car_plot}a and \ref{fig:car2car_plot}b), most of the red dots are grouped in a few clusters, indicating that most of the input images are translated into images with same azimuth, and that these models suffer from mode collapsing problem as predicted and shown in Figures \ref{fig:1d_illustration} and \ref{fig:toy}. Our proposed DiscoGAN (\ref{fig:car2car_plot}c), on the other hand, shows strong correlation between predicted angles of input and translated images, indicating that our model successfully discovers azimuth relation between the two domains. In this experiment, the translated images either have the same azimuth range (\ref{fig:car2car_plot}b), or the opposite (\ref{fig:car2car_plot}a and \ref{fig:car2car_plot}c) of the input images. %RMSE between the results and the groundtruth (blue lines) are also indicated.

%As shown in figure \ref{fig:car2car_plot}c, which plots the prediction results of the regressor for input and translated images for each network, out proposed DiscoGAN model indeed discovers azimuth as the relation between two domains and the red dots are . On the other hand, standard GAN model  

\begin{figure}[H]
  %\vskip 0.2in
  \begin{center}
  \centerline{\includegraphics[width=1.1\columnwidth]{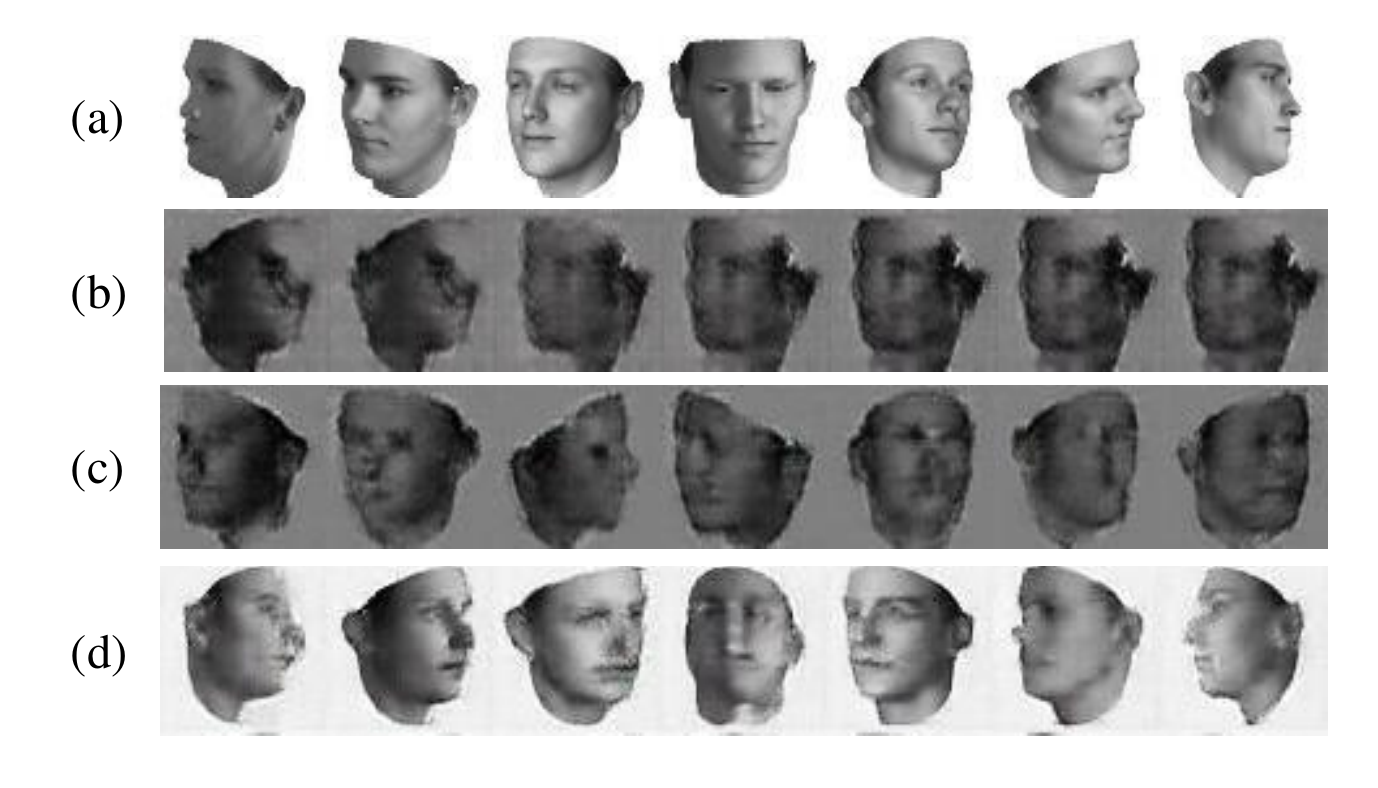}}
  \caption{Face to Face translation experiment. (a) input face images from -90$^\circ$ to +90$^\circ$ (b) results from a standard GAN (c) results from GAN with a reconstruction loss (d) results from our DiscoGAN. Here our model generated images in the opposite range, from +90$^\circ$ to -90$^\circ$.}
  \label{fig:face_toy}
  \end{center}
  \vskip -0.2in
  \end{figure}

Next, we use a Face dataset \cite{bfm09} shown in Figure \ref{fig:face_toy}a, in which the data images vary in azimuth rotation from -90$^\circ$ to +90$^\circ$. Similar to previous car to car experiment, input images in the -90$^\circ$ to +90$^\circ$ rotation range generated output images either in the same range, from -90$^\circ$ to +90$^\circ$, or the opposite range, from +90$^\circ$ to -90$^\circ$ when our proposed model was used (\ref{fig:face_toy}d). We also trained a standard GAN and a GAN with reconstruction loss for comparison. When a standard GAN and GAN with reconstruction loss were used, the generated images do not vary as much as the input images in terms of rotation. In this sense, similar to what has been shown in previous Car to Car experiment, the two models suffered from mode collapse.

\subsubsection{Face Conversion}
\begin{figure*}[htbp]
  %\vskip 0.2in
  \begin{center}
  \centerline{\includegraphics[width=\textwidth]{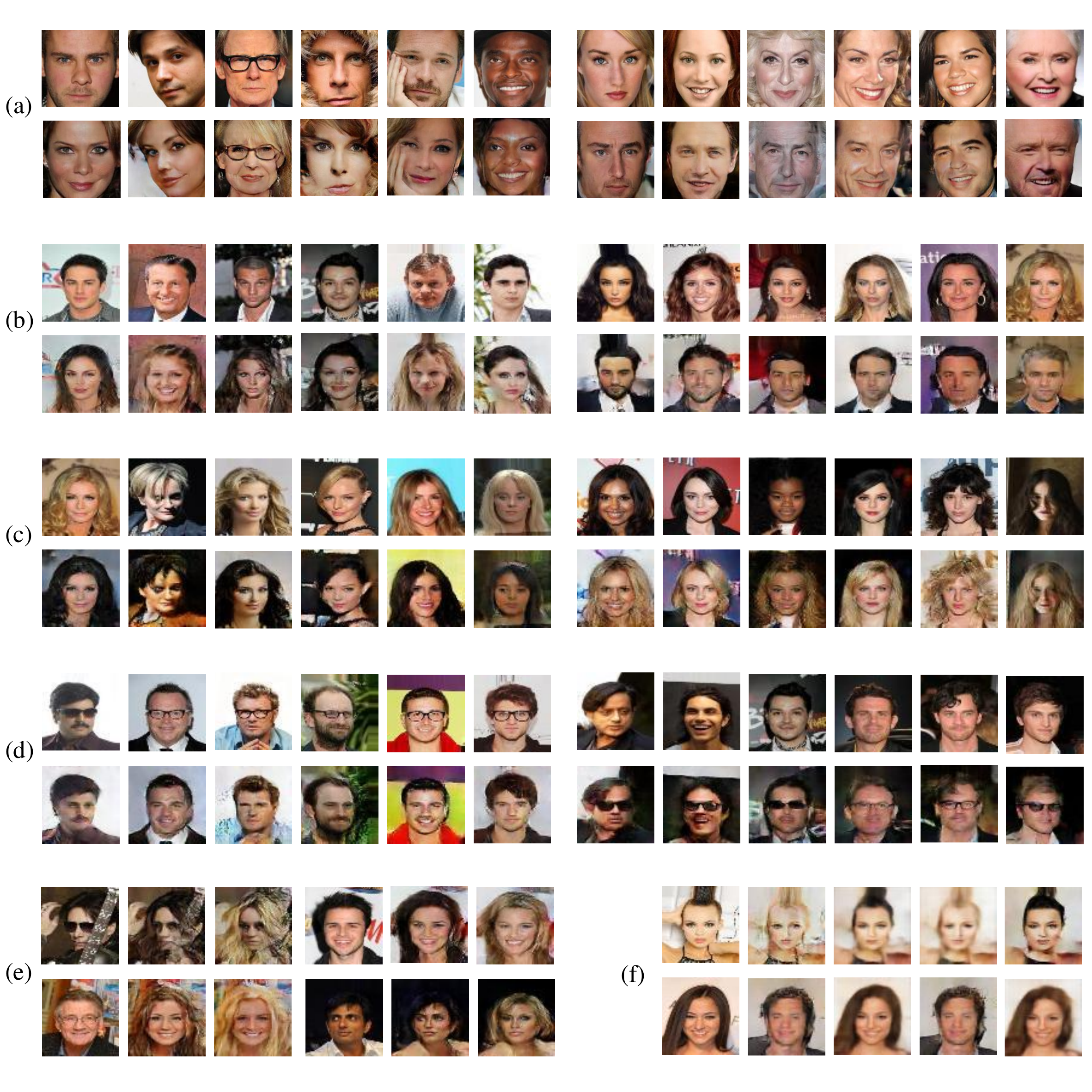}}
  \caption{(a,b) Translation of gender in Facescrub dataset and CelebA dataset.
  (c) Blond to black and black to blond hair color conversion in CelebA dataset.
  (d) Wearing eyeglasses conversion in CelebA dataset
  (e) Results of applying a sequence of conversion of gender and hair color (left to right)
  (f) Results of repeatedly applying the same conversions (upper: hair color, lower: gender)
  }
  \label{fig:face_trans}
  \end{center}
  \vskip -0.2in
  \end{figure*}

In terms of the amount of related information between two domains, we can consider a few extreme cases: two domains sharing almost all features and two domains sharing only one feature. To investigate former case, we applied the face attribute conversion task on CelebA dataset \cite{liu2015faceattributes}, where only one feature, such as gender or hair color, varies between two domains and all the other facial features are shared. The results are listed in Figure \ref{fig:face_trans}. 

In Figure \ref{fig:face_trans}a, we can see that various facial features are well-preserved while a single desired attribute (gender) is changed. Also, \ref{fig:face_trans}b and \ref{fig:face_trans}d shows that background is also well-preserved and images are visually natural, although the background does change in a few cases such as Figure \ref{fig:face_trans}c. An extension to this experiment was sequentially applying several translations -- for example, changing the gender and then the hair color (\ref{fig:face_trans}e), or repeatedly applying gender transforms (\ref{fig:face_trans}f).

    \begin{figure*}[htbp]
  %\vskip 0.2in
  \begin{center}
  \centerline{\includegraphics[width=\textwidth]{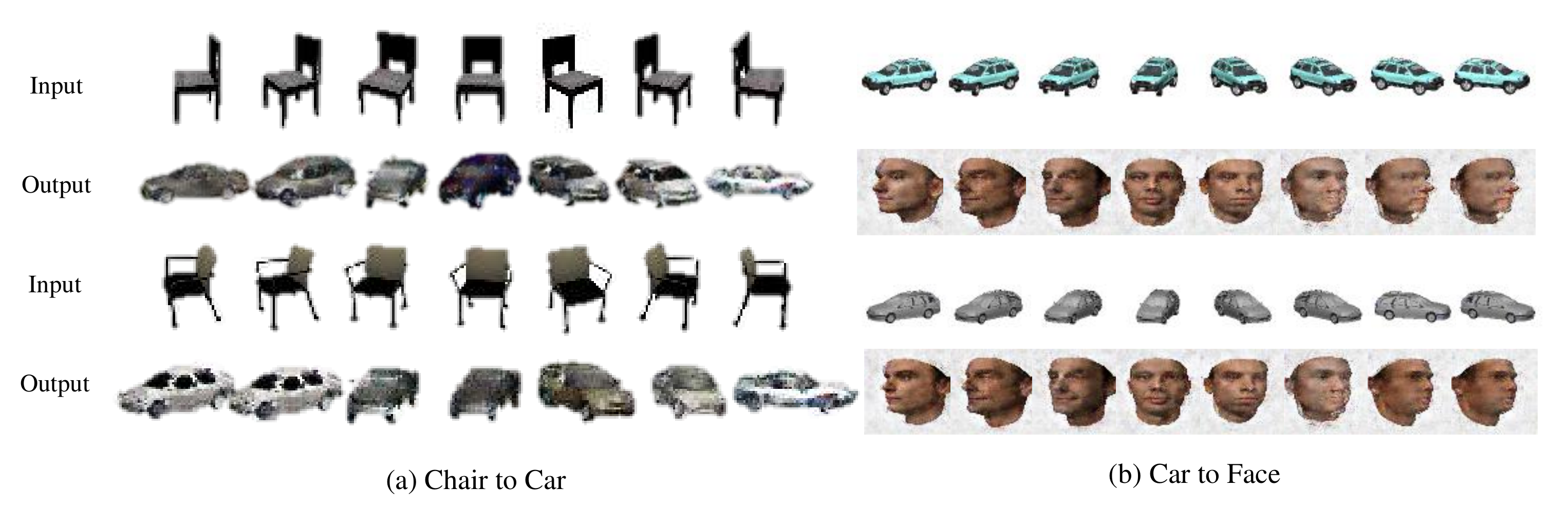}}
  \caption{Discovering relations of images from visually very different object classes. (a) chair to car translation. DiscoGAN is trained on chair and car images (b) car to face translation. DiscoGAN is trained on car and face images. Our model successfully pairs images with similar orientation. }
  \label{fig:exp_main_fig}
  \end{center}
  \vskip -0.2in
  \end{figure*}

  \begin{figure}[t]
  %\vskip 0.2in
  \begin{center}
  \centerline{\includegraphics[width=\columnwidth]{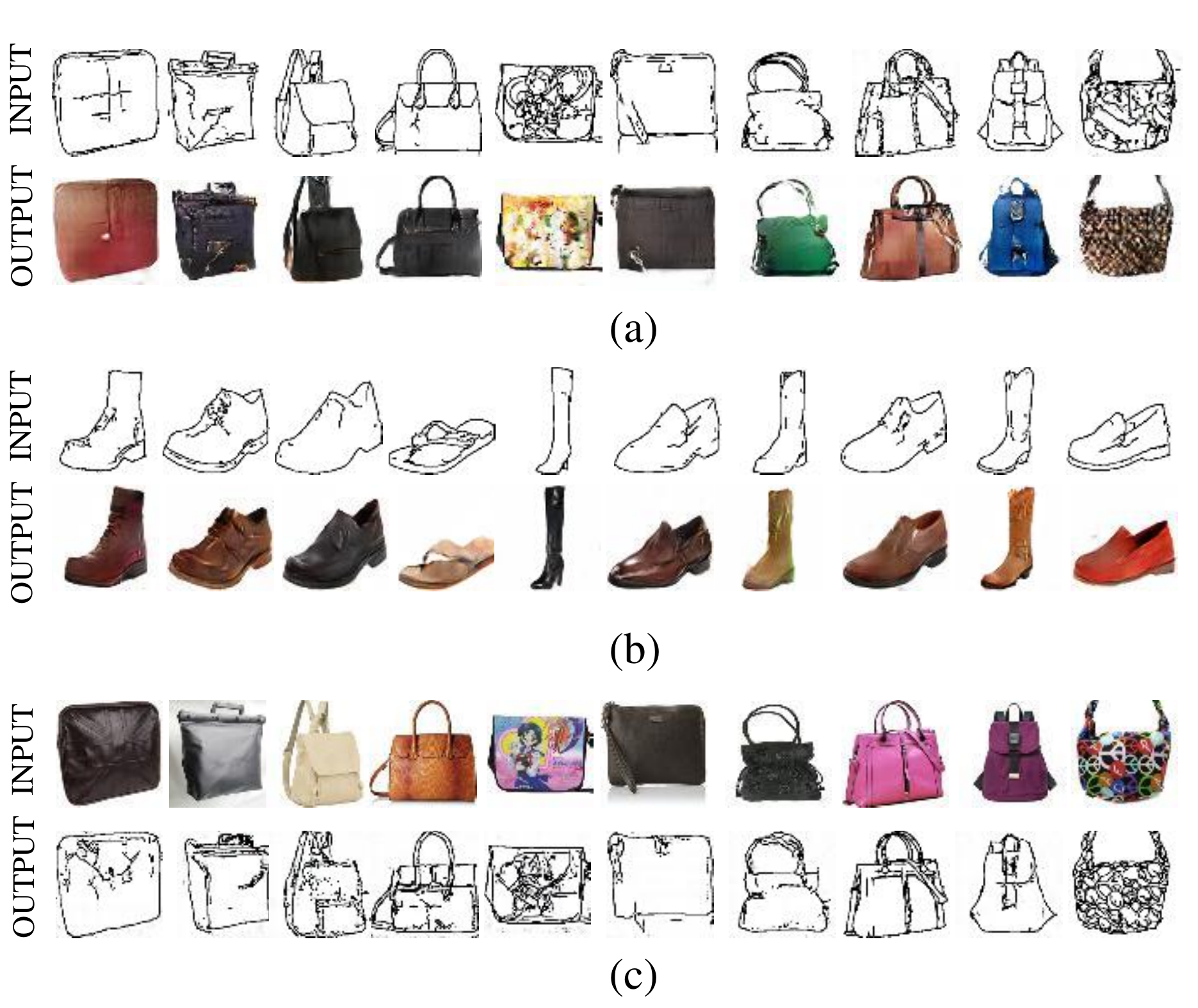}}
  \caption{ Edges to photos experiment. Our model is trained on a set of object sketches and colored images and learns to generate new sketches or photos. (a) colored images of handbags are generated from sketches of handbags, (b) colored images of shoes are generated from sketches of shoes, (c) sketches of handbags are generated from colored images of handbags
  }
  \label{fig:edge_to_photo}
  \end{center}
%  \vskip -0.2in
  \end{figure}

\subsubsection{Chair to Car, Car to Face}
We also investigated the opposite case where there is a single shared feature between two domains. 3D rendered images of chair \cite{aubry2014chair} and the previously used car and face datasets \cite{fidler2012car, bfm09} were used in this task. All three datasets vary along the azimuth rotation. Figure \ref{fig:exp_main_fig} shows the results of image-to-image translation from chair to car and from car to face datasets. The translated images clearly match the rotation feature of the input images while preserving visual features of car and face domain, respectively. 

\subsubsection{Edges-to-Photos}
Edges-to-photos is an interesting task as it is a 1-to-N problem, where a single edge image of items such as shoes and handbags can generate multiple colorized  images of such items. In fact, an edge image can be colored in infinitely many ways. We validated that our DiscoGAN performs very well on this type of image-to-image translation task and generate realistic photos of handbags \cite{zhu2016generative} and shoes \cite{fine-grained}. The generated images are presented in Figure \ref{fig:edge_to_photo}.

\subsubsection{Handbag to Shoes, Shoes to Handbag}
Finally, we investigated the case with two domains that are visually very different, where shared features are not explicit even to humans. We trained a DiscoGAN using previously used handbags and shoes datasets, not assuming any specific relation between those two. In the translation results shown in Figure \ref{fig:teaser}, our proposed model discovers fashion style as a related feature between the two domains. Note that translated results not only have similar colors and patterns, but they also have similar level of fashion formality as the input fashion item.

\section{Related Work}
\label{Related work}

Recently, a novel method to train generative models named Generative Adversarial Networks (GANs) \cite{goodfellow2014generative} was developed. A GAN is composed of two modules -- a generator $G$ and a discriminator $D$. The generator's objective is to generate (synthesize) data samples whose distribution closely matches that of real data samples while the discriminator's objective is to distinguish real ones from generated samples. The two models $G$ and $D$, formulated as a two-player minimax game, are trained simultaneously.

% \begin{equation} 
% \begin{split} 
% \min_G \max_D V(G,D) = \ &\mathbb{E}_{x_{\!_B} \sim P_{\!_B}}  \left[ \log D(x_{\!_B}) \right] + \\
%               \ &\mathbb{E}_{x_{\!_A} \sim P_{\!_A}} \left[ 1 - \log D(G(x_{\!_A}))\right] 
% \end{split} 
% \end{equation}
% Previous deep generative models often suffered from having many computationally intractable terms but GANs proposed a way around this problem and led to many novel works especially in the field of image generation. 

Researchers have studied GANs vigorously in two years: network models such as LAPGAN \cite{denton2015deep} and DCGAN \cite{radford2015unsupervised} and improved training techniques \cite{salimans2016improvedgans,arjovsky2017wgan}. More recent GAN works are described in \cite{goodfellow2016NIPStutorial}.    

%Many researchers have studied 

% Shortly after GANs were introduced, \citet{radford2015unsupervised} combined Convolutional Neural Nets (CNNs) with GANs and proposed Deep Convolutional Generative Adversarial Networks (DCGANs) for image generation. This approach bridged supervised learning of CNNs with unsupervised learning of GANs. In addition, \citet{denton2015deep} used a cascade of CNNs in a Laplacian pyramid framework to generate images in multiple steps. At each level of the cascade, a separate CNN that produces images in a coarse-to-fine manner is trained using the GAN approach. 
  
%\textbf{GANs for image generation (from some feature)} \\
%\subsection{Conditional GAN}
Several methods were developed to generate images based on GANs. Conditional Generative Adversarial Nets (cGANs) \cite{mirza2014conditional} use MNIST digit class label as an additional information to both generator and discriminator and can generate digit images of the specified class. Similarly, \citet{Dosovitskiy_2015_CVPR} showed that GAN can generate images of objects based on specified characteristic codes such as color and viewpoint. Other approaches used conditional features from a completely different domain for image generation. For example, \citet{reed2016generative} used encoded text description of images as the conditional information to generating images that match the description.

% VAE-GAN 
% MD-GAN
% 2-GANS 
% Unsupervised Domain Transfer  
%\textbf{GANs for Image-to-image translation (style transfer, colorization)} \\

% \citet{dong2017unsupervised} used a two-step unsupervised learning method to translate between different genders (``gender transformation'') or two entities (``face swap'') while preserving facial features.  

% \textbf{GAN+$\alpha$}\\
% There also have been works that augment ordinary GANs with additional networks. Bidirectional GAN (BiGAN) \cite{donahue2016adversarial} and Adversarially Learned Inference (ALI) \cite{dumoulin2016adversarially} independently proposed adding an additional encoder network that maps training examples in data space to latent space. This modification allows the model to learn feature representations in the latent space. 

%\textbf{Two GANs}\\
%\subsection{Multiple GANs}
Some researchers have attempted to use multiple GANs in prior works. \cite{liu2016coupled} proposed to couple two GANs (coupled generative adversarial networks, CoGAN) in which two generators and two discriminators are coupled by weight-sharing to learn the joint distribution of images in two different domains without using pair-wise data. In Stacked GANs (StackGAN) \cite{zhang2016stackgan}, two GANs are arranged sequentially where the Stage-I GAN generates low resolution images given text description and the Stage-II GAN improves the generated image into high resolution images. Similarly, Style and Structure GAN (S$^2$-GAN) \cite{wang2016s2gan} used two sequentially connected GANs where the Structure GAN first generates surface normal image and the Style GAN transforms it into natural indoor scene image.

In order to control specific attributes of an image, \citet{tejas2015dcign} proposed a method to disentangle specific factors by explicitly controlling target code. \citet{perarnau2016invertible} tackled image generation problems conditioned on specific attribute vectors by training an attribute predictor along with latent encoder.   
 
 %\cite{kingma14vae} for encoding latent vectors from input images.
%VAE-GAN \cite{larsen2015autoencoding} tackled the problem of image style transfer conditioned on specific latent vector by introducing a Variational Autoencoder
% \textbf{``Dual-learning'' using two GANs} \\
% An interesting approach to combining two GANs is using a dual-learning technique. In this learning technique, originally proposed in application to machine translation by \citet{he2016duallearningtranslation}, uses two tasks named the primal and the dual. The primal task, such as translating English-to-French, and the dual task, translating French-to-English, forms a closed loop in which the information signal from both models feedback through the loop to trains each other. Inspired by this work, \citet{shen2016duallearning} applied this method to training GANs for image attribute manipulation. We utilize a similar dual-learning approach, but instead of manipulating images in the same domain, we translate images between two different domains as the primal and dual tasks.
%\subsection{Image-to-Image Translation}
In addition to using conditional information such as class labels and text encodings, several works in the field of image-to-image translation used images of one domain to generate images in another domain. \cite{isola2016image} translated black-and-white images to colored images by training on paired black-and-white and colored image data. Similarly, \citet{taigman2016unsupervised} translated face images to emojis by providing image features from pre-trained face recognition module as conditional input to a GAN. 

Recently, \citet{tche2017mdgan} tackled mode-collapsing and instability problems in GAN training. They introduced two ways of regularizing general GAN objective -- geometric metrics regularizer and mode regularizer. 
%\textbf{References yet to be inserted} \\
%Mode Regularized Generative Adversarial Networks \cite{che2016mode} \hsk{저희가 double model을 쓰는 이유를 어떻게 풀어낼지에 따라 MD-GAN이랑 연결고리가 나올것 같네요. 일단 이부분은 보류. mode regularizer? missing mode problem?}
%Invertible Conditional GANs for Image Editing %\cite{perarnau2016invertible} \\
%Autoencoding Beyond Pixels using a Learned Similarity Metric \cite{larsen2015autoencoding} \jiwon{여기 보면 CelebA가지고 Visual Attribute Vector 실험하는게 많이 나오는데, 이미 참고하지 않았다면 보면 괜찮을 것 같네요. Section 4.1.1 등}
\section{Conclusion}
This paper presents a learning method to discover cross-domain relations with a generative adversarial network called DiscoGAN. Our approach works without any explicit pair labels and learns to relate datasets from very different domains. We have demonstrated that DiscoGAN can generate  high-quality images with transferred style. One possible future direction is to modify DiscoGAN to handle mixed modalities (e.g. text and image).

\bibliographystyle{icml2017}
\bibliography{pairgan}

%  \begin{figure}[htbp]
%   %\vskip 0.2in
%   \begin{center}
%   \centerline{\includegraphics[width=\columnwidth]{Figures/Result_car2car_angle_version_v0_0_2_2017_02_24}}
%   \caption{car to car angle version}
%   \label{fig:car_to_car_angle}
%   \end{center}
%   \vskip -0.2in
%   \end{figure}

%   \begin{figure}[htbp]
%   %\vskip 0.2in
%   \begin{center}
%   \centerline{\includegraphics[width=\columnwidth]{Figures/Result_edge_to_shoes_by_iterations_v_0_0_1_2017_02_24}}
%   \caption{ Edge to shoes by iteration
%   }
%   \label{fig:edge_to_shoes}
%   \end{center}
%   \vskip -0.2in
%   \end{figure}

\end{document}